\documentclass[11pt]{article}
\usepackage{acl2016}
\usepackage{times}
\usepackage{url}
\usepackage{latexsym}
\usepackage{graphicx}
\usepackage{mathtools}

\aclfinalcopy 
\def\aclpaperid{312} 

\title{Sentence Rewriting for Semantic Parsing}

\author{Bo Chen \ \ \ Le Sun \ \ \ Xianpei Han \ \ \ Bo An \\
  State Key Laboratory of Computer Sciences \\
  Institute of Software, Chinese Academy of Sciences, China. \\
  {\tt \{chenbo, sunle, xianpei, anbo\}@iscas.ac.cn} \\
  }

\date{}

\begin{document}
\maketitle
\begin{abstract}
  A major challenge of semantic parsing is the vocabulary mismatch problem between
  natural language and target ontology.
  In this paper, we propose a sentence rewriting based semantic parsing method,
   which can effectively resolve the mismatch problem by rewriting a sentence
    into a new form which has the same structure with its target logical form.
     Specifically, we propose two sentence-rewriting methods for two common types of mismatch:
      a dictionary-based method for 1-N mismatch and a template-based method for N-1 mismatch.
      We evaluate our sentence rewriting based semantic parser on the benchmark semantic parsing dataset -- WEBQUESTIONS.
        Experimental results show that our system outperforms the base system with a 3.4\% gain in F1,
         and generates logical forms more accurately and parses sentences more robustly.
\end{abstract}

\section{Introduction}
Semantic parsing is the task of mapping natural language sentences
into logical forms which can be executed on a knowledge base ~\cite{zelle:aaai96,DBLP:conf/uai/ZettlemoyerC05,kate-mooney:2006:COLACL,wong-mooney:2007:ACLMain,lu-EtAl:2008:EMNLP,kwiatkowksi-EtAl:2010:EMNLP}. Figure 1 shows an example of semantic parsing.
 Semantic parsing is a fundamental technique of natural language understanding,
 and has been used in many applications, such as question answering ~\cite{liang-jordan-klein:2011:ACL-HLT2011,he-EtAl:2014:EMNLP20142,DBLP:conf/aaai/ZhangHL016}
  and information extraction ~\cite{krishnamurthy-mitchell:2012:EMNLP-CoNLL,choi-kwiatkowski-zettlemoyer:2015:ACL-IJCNLP,parikh-poon-toutanova:2015:NAACL-HLT}.

  \begin{figure}
     \centering
     \includegraphics[width=0.5\textwidth]{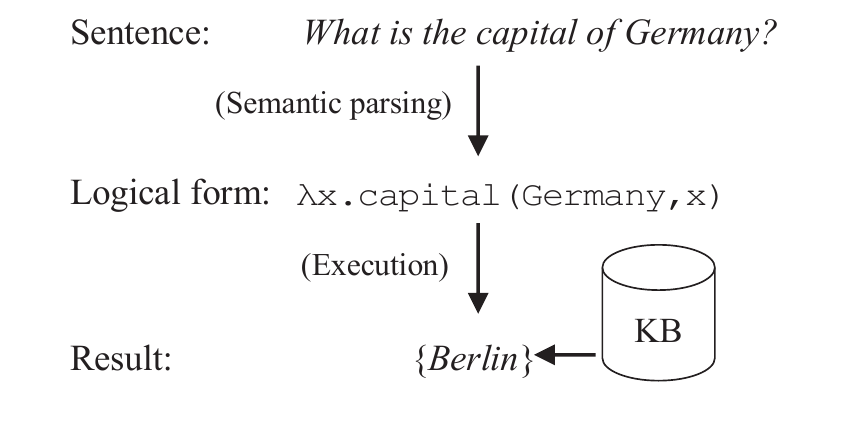}
     \caption{An example of semantic parsing.}
     \label{example-figure}
 \end{figure}

  Semantic parsing, however, is a challenging task.
   Due to the variety of natural language expressions,
   the same meaning can be expressed using different sentences.
    Furthermore, because logical forms depend on the vocabulary of target-ontology,
    a sentence will be parsed into different logical forms when using different ontologies.
    For example, in below the two sentences $s_1$ and $s_2$ express the same meaning,
    and they both can be parsed into the two different logical forms  $lf_1$ and  $lf_2$
    using different ontologies.

    \begin{table}[h]
    \small
    \centering
    \begin{tabular}{ll}
    $s_1$ & \textit{What is the population of Berlin?} \\
    $s_2$ &  \textit{How many people live in Berlin?} \\
    $lf_1$ & \texttt{$\lambda$x.population(Berlin,x)} \\
    $lf_2$ & \texttt{count($\lambda$x.person(x)$\wedge$live(x,Berlin))}  \\
    \end{tabular}
    \end{table}

    Based on the above observations,
    one major challenge of semantic parsing is the structural mismatch
    between a natural language sentence and its target logical form,
    which are mainly raised by the vocabulary mismatch between natural language and ontologies.
    Intuitively, if a sentence has the same structure with its target logical form,
    it is easy to get the correct parse, e.g., a semantic parser can easily parse
    $s_1$ into $lf_1$ and $s_2$ into $lf_2$. On the contrary,
    it is difficult to parse a sentence into its logic form when they have different structures,
    e.g., $s_1 \rightarrow lf_2$ or $s_2 \rightarrow lf_1$.

    To resolve the vocabulary mismatch problem,
    this paper proposes a sentence rewriting approach for semantic parsing,
     which can rewrite a sentence into a form which will have the same structure
     with its target logical form.
     Table 1 gives an example of our rewriting-based semantic parsing method.
     In this example, instead of parsing the sentence
     ``\textit{What is the name of Sonia Gandhi’s daughter?}'' into its structurally different
     logical form \texttt{childOf.S.G.$\wedge$gender.female} directly, our method will first rewrite the sentence into the form
     ``\textit{What is the name of Sonia Gandhi’s female child?}'',
     which has the same structure with its logical form,
     then our method will get the logical form by parsing this new form.
     In this way, the semantic parser can get the correct parse more easily.
     For example, the parse obtained through traditional method will result in the wrong answer
     ``\textit{Rahul Gandhi}'', because it cannot identify the vocabulary mismatch between ``\textit{daughter}'' and \texttt{child$\wedge$female\footnote{In this paper, we may simplify logical forms for readability, e.g., \texttt{female} for \texttt{gender.female}.}}.
     By contrast, by rewriting ``\textit{daughter}'' into ``\textit{female child}'',
     our method can resolve this vocabulary mismatch.

\begin{table}
 \small
     (a)  An example using traditional method \\
     \begin{tabular}{|p{205pt}|}
       \hline
       \parbox{205pt}{$s_0:$ \textit{What is the name of Sonia Gandhi’s daughter?}
      } \\
      \parbox{205pt}{ $l_0\ :$ \texttt{$\lambda$x.child(S.G.,x)}
     } \\
      \parbox{205pt}{$r_0:$ \textit{\{Rahul Gandhi (Wrong answer), Priyanka Vadra\}}
    } \\
    \hline
     \end{tabular}
     (b)  An example using our method \\
     \begin{tabular}{|p{205pt}|}
       \hline
        \parbox{205pt}{$s_0:$ \textit{What is the name of Sonia Gandhi’s daughter?}
      } \\
       \parbox{205pt}{$s_1:$ \textit{What is the name of Sonia Gandhi’s female child?}
     } \\
       \parbox{205pt}{ $l_1\ :$ \texttt{$\lambda$x.child(S.G.,x)$\wedge$gender(x,female)}
     } \\
     \parbox{205pt}{$r_1:$ \textit{\{Priyanka Vadra\}}
    } \\
    \hline
     \end{tabular}
     \caption{\label{simple-example-table} Examples of (a) sentences $s_0$, possible logical form $l_0$
     from traditional semantic parser, result $r_0$ for the logical form $l_0$;
     (b) possible sentence $s_1$ from rewriting for the original sentence $s_0$,
     possible logical form $l_1$ for sentence $s_1$, result $r_1$ for $l_1$.
     \textit{Rahul Gandhi} is a wrong answer, as he is the son of \textit{Sonia Gandhi}.}
\end{table}

     Specifically, we identify two common types of vocabulary mismatch in semantic parsing:
  \begin{enumerate}
    \item 	1-N mismatch: a simple word may correspond to a compound formula.
     For example, the word ``\textit{daughter}'' may correspond to the compound formula \texttt{child$\wedge$female}.

     \item N-1 mismatch: a logical constant may correspond to a complicated
     natural language expression, e.g., the formula \texttt{population} can be expressed
     using many phrases such as ``\textit{how many people}'' and ``\textit{live in}''.

    \end{enumerate}

    To resolve the above two vocabulary mismatch problems,
    this paper proposes two sentence rewriting algorithms:
    One is a dictionary-based sentence rewriting algorithm,
    which can resolve the 1-N mismatch problem by rewriting a word using its explanation
    in a dictionary. The other is a template-based sentence rewriting algorithm,
    which can resolve the N-1 mismatch problem by rewriting complicated expressions
    using paraphrase template pairs.

    Given the generated rewritings of a sentence,
    we propose a ranking function to jointly choose the optimal
    rewriting and the correct logical form, by taking both the rewriting features
    and the semantic parsing features into consideration.

    We conduct experiments on the benchmark WEBQUESTIONS dataset ~\cite{berant-EtAl:2013:EMNLP}.
     Experimental results show that our method can effectively resolve the vocabulary mismatch problem
     and achieve accurate and robust performance.

     The rest of this paper is organized as follows.
     Section 2 reviews related work. Section 3 describes our sentence rewriting method
     for semantic parsing. Section 4 presents the scoring function which can jointly ranks rewritings
     and logical forms. Section 5 discusses experimental results. Section 6 concludes this paper.

\section{Related Work}   
Semantic parsing has attracted considerable research attention in recent years.
Generally, semantic parsing methods can be categorized into synchronous context free grammars (SCFG)
based methods ~\cite{wong-mooney:2007:ACLMain,TACL654,li-EtAl:2015:EMNLP3},
syntactic structure based methods ~\cite{ge-mooney:2009:ACLIJCNLP,reddy_largescale_2014,reddy_transforming_2016},
combinatory categorical grammars (CCG) based methods
~\cite{zettlemoyer-collins:2007:EMNLP-CoNLL2007,kwiatkowksi-EtAl:2010:EMNLP,kwiatkowski-EtAl:2011:EMNLP,krishnamurthy-mitchell:2014:P14-1,wang-kwiatkowski-zettlemoyer:2014:EMNLP2014,artzi-lee-zettlemoyer:2015:EMNLP},
and dependency-based compositional semantics (DCS) based methods
~\cite{liang-jordan-klein:2011:ACL-HLT2011,berant-EtAl:2013:EMNLP,berant-liang:2014:P14-1,TACL646,pasupat2015compositional,wang-berant-liang:2015:ACL-IJCNLP}.

One major challenge of semantic parsing is how to scale to open-domain situation
like Freebase and Web. A possible solution is to learn lexicons from large amount of web text
and a knowledge base using a distant supervised method ~\cite{krishnamurthy-mitchell:2012:EMNLP-CoNLL,cai-yates:2013:ACL2013,berant-EtAl:2013:EMNLP}.
Another challenge is how to alleviate the burden of annotation.
 A possible solution is to employ distant-supervised techniques
 ~\cite{clarke-EtAl:2010:CONLL,liang-jordan-klein:2011:ACL-HLT2011,cai-yates:2013:*SEM,artzi-zettlemoyer:2013:TACL},
 or unsupervised techniques ~\cite{poon-domingos:2009:EMNLP,goldwasser-EtAl:2011:ACL-HLT2011,poon:2013:ACL2013}.

 There were also several approaches focused on the mismatch problem.
Kwiatkowski et al. \shortcite{kwiatkowski-EtAl:2013:EMNLP} addressed the ontology mismatch problem
 (i.e., two ontologies using different vocabularies)
 by first parsing a sentence into a domain-independent underspecified logical form,
 and then using an ontology matching model to transform this underspecified logical form
 to the target ontology. However, their method is still hard to deal with the 1-N and the N-1 mismatch
  problems between natural language and target ontologies.
Berant and Liang \shortcite{berant-liang:2014:P14-1} addressed the structure mismatch problem between natural language
  and ontology by generating a set of canonical utterances for each candidate logical form,
  and then using a paraphrasing model to rerank the candidate logical forms.
  Their method addresses mismatch problem in the reranking stage,
  cannot resolve the mismatch problem when constructing candidate logical forms.
  Compared with these two methods, we approach the mismatch problem in the parsing stage,
  which can greatly reduce the difficulty of constructing the correct logical form,
  through rewriting sentences into the forms which will be structurally consistent with
  their target logic forms.

  Sentence rewriting (or paraphrase generation) is the task of generating new sentences
  that have the same meaning as the original one. Sentence rewriting has been used in
  many different tasks, e.g., used in statistical machine translation to resolve
  the word order mismatch problem ~\cite{collins-koehn-kucerova:2005:ACL,he-EtAl:2015:EMNLP}.
  To our best knowledge, this paper is the first work to apply sentence rewriting for vocabulary mismatch problem in semantic parsing.

\section{Sentence Rewriting for Semantic Parsing}   

As discussed before, the vocabulary mismatch between natural language
and target ontology is a big challenge in semantic parsing.
In this section, we describe our sentence rewriting algorithm for solving the mismatch problem.
Specifically, we solve the 1-N mismatch problem by dictionary-based rewriting
and solve the N-1 mismatch problem by template-based rewriting. The details are as follows.

\subsection{Dictionary-based Rewriting}   

In the 1-N mismatch case, a word will correspond to a compound formula,
e.g., the target logical form of the word ``\textit{daughter}'' is \texttt{child$\wedge$female} (Table 2 has more examples).

To resolve the 1-N mismatch problem, we rewrite the original word (``\textit{daughter}'') into an expression
(``\textit{female child}'') which will have the same structure with its target logical form (\texttt{child$\wedge$female}).
In this paper, we rewrite words using their explanations in a dictionary.
This is because each word in a dictionary will be defined by a detailed explanation using simple words,
which often will have the same structure with its target formula.
Table 2 shows how the vocabulary mismatch between a word and its logical form can be resolved
using its dictionary explanation. For instance, the word ``\textit{daughter}'' is explained as ``\textit{female child}''
in Wiktionary, which has the same structure as \texttt{child$\wedge$female}.

 \begin{table}
  \small
 \begin{tabular}{|p{46pt}|p{69pt}|p{65pt}|}
 \hline
 \parbox{46pt}{\centering\textbf{Word}
 } & \parbox{69pt}{\centering\textbf{Logical Form}
} & \parbox{65pt}{\centering\textbf{Wiktionary Explanation}
 } \\
 \hline
 \parbox{46pt}{\centering\textit{son}
 } & \parbox{69pt}{\centering\texttt{child$\wedge$male}
 } & \parbox{65pt}{\centering\textit{male child}
 } \\
 \hline
 \parbox{46pt}{\centering\textit{actress}
 } & \parbox{69pt}{\centering\texttt{actor$\wedge$female}
 } & \parbox{65pt}{\centering\textit{female actor}
 } \\
 \hline
 \parbox{46pt}{\centering\textit{father}
 } & \parbox{69pt}{\centering\texttt{parent$\wedge$male}
 } & \parbox{65pt}{\centering\textit{male parent}
 } \\
 \hline
 \parbox{46pt}{\centering\textit{grandaprent}
 } & \parbox{69pt}{\centering\texttt{parent$\wedge$parent}
 } & \parbox{65pt}{\centering\textit{parent of one's parent}
 } \\
 \hline
 \parbox{46pt}{\centering\textit{brother}
 } & \parbox{69pt}{\centering\texttt{sibling$\wedge$male}
 } & \parbox{65pt}{\centering\textit{male sibling}
 } \\
 \hline
 \end{tabular}
 \caption{\label{wik-table}Several examples of words,
  their logical forms and their explanations in Wiktionary.}
 \end{table}

In most cases, only common nouns will result in the 1-N mismatch problem.
Therefore, in order to control the size of rewritings,
this paper only rewrite the common nouns in a sentence by replacing them with
their dictionary explanations. Because a sentence usually will not contain too many common nouns,
the size of candidate rewritings is thus controllable.
Given the generated rewritings of a sentence, we propose a sentence selection model
to choose the best rewriting using multiple features (See details in Section 4).

Table 3 shows an example of the dictionary-based rewriting.
In Table 3, the example sentence $s$ contains two common nouns (``\textit{name}'' and ``\textit{daughter}''),
therefore we will generate three rewritings $r_1$, $r_2$ and $r_3$. Among these rewritings,
the candidate rewriting $r_2$ is what we expected, as it has the same structure
with the target logical form and doesn't bring extra noise
(i.e., replacing ``\textit{name}'' with its explanation ``\textit{reputation}'').

\begin{table}
\small
\centering
\begin{tabular}{|p{205pt}|}
\hline
\parbox{205pt}{\ $s$\ : \textit{What is the name of Sonia Gandhi’s daughter?}} \\ \hline
\parbox{205pt}{$r_1$: \textit{What is the reputation of Sonia Gandhi’s daughter?}} \\ \hline
\parbox{205pt}{$r_2$: \textit{What is the name of Sonia Gandhi’s female child?}} \\ \hline
\parbox{205pt}{$r_3$: \textit{What is the reputation of Sonia Gandhi’s female child?}} \\
\hline
\end{tabular}
\caption{\label{example-table}An example of the dictionary-based sentence rewriting.}
\end{table}

For the dictionary used in rewriting, this paper uses Wiktionary.
Specifically, given a word, we use its ``Translations'' part in the Wiktionary as its explanation.
Because most of the 1-N mismatch are caused by common nouns, we only collect the explanations of common nouns.
 Furthermore, for polysomic words which have several explanations,
 we only use their most common explanations. Besides, we ignore explanations whose length are longer than 5.

\subsection{Template-based Rewriting}    

In the N-1 mismatch case, a complicated natural language expression will be mapped
to a single logical constant. For example, considering the following mapping from
the natural language sentence $s$ to its logical form $lf$ based on Freebase ontology:

\begin{table}[h]
\centering
\begin{tabular}{c}
$s$: \textit{How many people live in Berlin?} \\
$lf$:  \texttt{$\lambda$x.population(Berlin,x)}  \\
\end{tabular}
\end{table}

\noindent where the three words: ``\textit{how many}'' (\texttt{count}), ``\textit{people}'' (\texttt{people}) and ``\textit{live in}'' (\texttt{live})
will map to the predicate \texttt{population} together. Table 4 shows more N-1 examples.

\begin{table}[h]
\small
\centering
\begin{tabular}{|p{125pt}|p{70pt}|}
  \hline
  \parbox{125pt}{\centering\textbf{Expression}
  } & \parbox{70pt}{\centering\textbf{Logical constant}
 } \\
 \hline
 \parbox{125pt}{\centering\textit{how many, people, live in}
 } & \parbox{70pt}{\centering\texttt{population}
 } \\
 \hline
 \parbox{125pt}{\centering\textit{how many, people, visit, annually}
 } & \parbox{70pt}{\centering\texttt{annual-visit}
  } \\
 \hline
 \parbox{125pt}{\centering\textit{what money, use}
 } & \parbox{70pt}{\centering\texttt{currency}
 } \\
 \hline
 \parbox{125pt}{\centering\textit{what school, go to}
 } & \parbox{70pt}{\centering\texttt{education}
 } \\
 \hline
 \parbox{125pt}{\centering\textit{what language, speak, officially}
 } & \parbox{70pt}{\centering\texttt{official-\\language}
  } \\
 \hline
\end{tabular}
\caption{\label{mismatch-table} Several N-1 mismatch examples.}
\end{table}

To resolve the N-1 mismatch problem, we propose a template rewriting algorithm,
which can rewrite a complicated expression into its simpler form. Specifically,
we rewrite sentences based on a set of paraphrase template pairs $P = \{(t_{i1},t_{i2}) | i=1,2,...,n\}$,
where each template $t$ is a sentence with an argument slot \$y, and $t_{i1}$ and $t_{i2}$ are paraphrases.
In this paper, we only consider single-slot templates. Table 5 shows several paraphrase template pairs.

\begin{table}
\small
\centering
\begin{tabular}{|p{100pt}|p{95pt}|}
  \hline
  \parbox{100pt}{\centering\textbf{Template 1}
  } & \parbox{95pt}{\centering\textbf{Template 2}
 } \\
 \hline
 \parbox{100pt}{\centering\textit{How many people live in \$y}
 } & \parbox{95pt}{\centering\textit{What is the population of \$y}
 } \\
 \hline
 \parbox{100pt}{\centering\textit{What money in \$y is used}
 } & \parbox{95pt}{\centering\textit{What is the currency of \$y}
  } \\
 \hline
 \parbox{100pt}{\centering\textit{What school did \$y go to}
 } & \parbox{95pt}{\centering\textit{What is the education of \$y}
 } \\
 \hline
 \parbox{100pt}{\centering\textit{What language does \$y speak officially }
 } & \parbox{95pt}{\centering\textit{What is the official language of \$y  }
 } \\
 \hline
\end{tabular}
\caption{\label{paraphrase-template-table} Several examples of paraphrase template pairs.}
\end{table}

\noindent Given the template pair database and a sentence, our template-based rewriting algorithm works as follows:

\begin{enumerate}
  \item Firstly, we generate a set of candidate templates $ST = \{st_1, st_2,...,st_n\}$ of
   the sentence by replacing each named entity within it by ``\$y''. For example, we will
   generate template ``\textit{How many people live in \$y}'' from the sentence ``\textit{How many people live in Berlin}''.
  \item 	Secondly, using the paraphrase template pair database, we retrieve all possible
  rewriting template pairs $(t_1,t_2)$ with $t_1 \in ST$, e.g., we can retrieve template pair
  (``\textit{How many people live there in \$y}'', ``\textit{What is the population of \$y}'' for $t_2$) using the above $ST$.
  \item 	Finally, we get the rewritings by replacing the argument slot ``\$y'' in template $t_2$
  with the corresponding named entity. For example, we get a new candidate sentence ``\textit{What is the population of Berlin}''
  by replacing ``\$y'' in $t_2$ with Berlin. In this way we can get the rewriting we expected,
  since this rewriting will match its target logical form \texttt{population(Berlin)}.
\end{enumerate}

To control the size and measure the quality of rewritings using
a specific template pair, we also define several features and the similarity between template pairs
(See Section 4 for details).

\begin{table}
\small
\centering
\begin{tabular}{|p{195pt}|}
  \hline
  \parbox{195pt}{\textit{How many people live in chembakolli?}
  } \\
  \parbox{195pt}{\textit{How many people is in chembakolli?}
  } \\
  \parbox{195pt}{\textit{How many people live in chembakolli india?}
  } \\
  \parbox{195pt}{\textit{How many people live there chembakolli?}
  } \\
  \parbox{195pt}{\textit{How many people live there in chembakolli?}
  } \\
  \parbox{195pt}{\textit{What is the population of Chembakolli india?}
  } \\
  \hline
  \parbox{195pt}{\textit{What currency is used on St Lucia?}
  } \\
  \parbox{195pt}{\textit{What is st lucia money?}
  } \\
  \parbox{195pt}{\textit{What is the money used in st lucia?}
  } \\
  \parbox{195pt}{\textit{What kind of money did st lucia have?}
  } \\
  \parbox{195pt}{\textit{What money do st Lucia use?}
  } \\
  \parbox{195pt}{\textit{Which money is used in St Lucia?}
  } \\
  \hline
\end{tabular}
\caption{\label{clusters-table} Two paraphrase clusters from the WikiAnswers corpus.}
\end{table}

To build the paraphrase template pair database, we employ the method described in Fader et al. \shortcite{Fader:2014:OQA:2623330.2623677}
 to automatically collect paraphrase template pairs. Specifically, we use the WikiAnswers paraphrase corpus
 ~\cite{fader-zettlemoyer-etzioni:2013:ACL2013}, which contains 23 million question-clusters,
 and all questions in the same cluster express the same meaning. Table 6 shows two paraphrase clusters
 from the WikiAnswers corpus. To build paraphrase template pairs, we first replace the shared noun words
 in each cluster with the placeholder ``\$y'', then each two templates in a cluster will form a paraphrase
 template pair. To filter out noisy template pairs, we only retain salient paraphrase template pairs
 whose co-occurrence count is larger than 3.

\section{Sentence Rewriting based Semantic Parsing}   

In this section we describe our semantic rewriting based semantic parsing system.
Figure 2 presents the framework of our system. Given a sentence, we first rewrite it into a set of
new sentences, then we generate candidate logical forms for each new sentence using a base semantic parser,
 finally we score all logical forms using a scoring function and output the best logical form as
 the final result. In following, we first introduce the used base semantic parser,
 then we describe the proposed scoring function.

\subsection{Base Semantic Parser}   

In this paper, we produce logical forms for each sentence rewritings using an
agenda-based semantic parser ~\cite{TACL646}, which is based on the lambda-DCS
proposed by Liang \shortcite{liang2013lambda}. For parsing, we use the lexicons and the grammars released by Berant et al. \shortcite{berant-EtAl:2013:EMNLP},
 where lexicons are used to trigger unary and binary predicates, and grammars are used to conduct logical forms.
  The only difference is that we also use the composition rule to make the parser can handle complicated questions
   involving two binary predicates, e.g., \texttt{child.obama$\wedge$gender.female}.

   \begin{figure}
      \centering
      \includegraphics[width=0.5\textwidth]{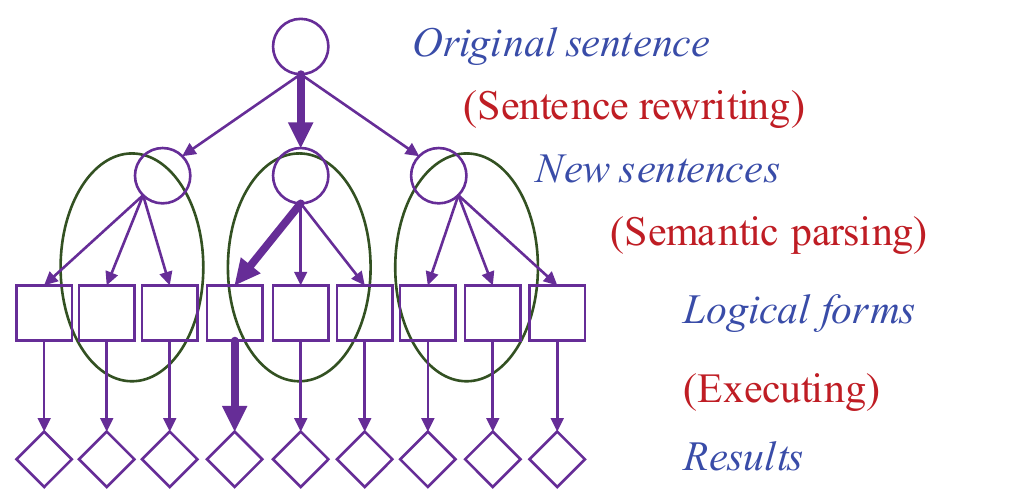}
      \caption{The framework of our sentence rewriting based semantic parsing.}
      \label{framework-figure}
   \end{figure}

  For model learning and sentence parsing, the base semantic parser learned a scoring function
  by modeling the policy as a log-linear distribution over (partial) agenda derivations Q:

  \begin{equation}
  p_{\theta}(a|s)=\frac{\exp\{\phi(a)^T\theta)\}}{\sum_{a'\in A}\exp\{\phi(a')^T\theta)\}}
  \end{equation}

  \noindent The policy parameters are updated as follows:

  \begin{equation}
  \theta\leftarrow\theta+\eta R(h_{target})\sum\nolimits_{t=1}^T\delta(h_{target})
  \end{equation}
  \begin{equation}
    \label{wave kinematic}
    \begin{split}
        \delta_t(h) &= \nabla_{\theta}\log p_{\theta}(a_t|s_t)  \\
        &=\phi(a_t)-E_{p_{\theta}(a_t'|s_t)}[\phi(a_t')]
    \end{split}
  \end{equation}

  \noindent The reward function $R(h)$ measures the compatibility of the resulting derivation,
  and $\eta$ is the learning rate which is set using the AdaGrad algorithm ~\cite{Duchi:2011:ASM:1953048.2021068}.
  The target history $h_{target}$ is generated from the root derivation $d^*$ with highest reward out of
  the $K$ (beam size) root derivations, using local reweighting and history compression.

\subsection{Scoring Function}   

To select the best semantic parse, we propose a scoring function which can take both sentence
rewriting features and semantic parsing features into consideration. Given a sentence $x$,
a generated rewriting $x'$ and the derivation $d$ of $x'$, we score them using follow function:

\begin{equation*}
  \label{wave kinematic}
  \begin{split}
    score(x,x',d) &=\theta\cdot\phi(x,x',d)   \\
    &= \theta_1\cdot\phi(x,x')+\theta_2\cdot\phi(x',d)
  \end{split}
\end{equation*}

This scoring function is decomposed into two parts: one for sentence rewriting -- $\theta_1\cdot\phi(x,x')$
and the other for semantic parsing -- $\theta_2\cdot\phi(x',d)$. Following Berant and Liang \shortcite{TACL646}, we update the
parameters $\theta_2$ of semantic parsing features as the same as (2). Similarly,
the parameters $\theta_1$ of sentence rewriting features are updated as follows:

\begin{equation*}
  \theta_1\leftarrow\theta_1+\eta R(h_{target}^*)\delta(x,x'^*)
\end{equation*}
\begin{equation*}
  \label{wave kinematic}
  \begin{split}
    \delta(x,x'^*)&=\nabla \log p_{\theta_1}(x'^*|x)  \\
    &=\phi(x,x'^*)-E_{p_{\theta_1}(x'|x)}[\phi(x,x')]
  \end{split}
\end{equation*}

\noindent where the learning rate $\eta$ is set using the same algorithm in Formula (2).

\subsection{Parameter Learning Algorithm}  

To estimate the parameters $\theta_1$ and $\theta_2$, our learning algorithm uses a set of
question-answer pairs $(x_i,y_i)$. Following Berant and Liang \shortcite{TACL646}, our updates for $\theta_1$ and $\theta_2$ do not maximize reward nor the log-likelihood.
 However, the reward provides a way to modulate the magnitude of the updates. Specifically,
 after each update, our model results in making the derivation, which has the highest reward,
 to get a bigger score. Table 7 presents our learning algorithm.

 \begin{table}
 \centering
 \small
 \begin{tabular}{p{205pt}}
 \hline
  \parbox{205pt}{\textbf{Input}: Q/A pairs $\{(x_i,y_i):i=1...n\}$; Knowledge base $\mathcal{K}$;
   Number of sentences $N$; Number of iterations $T$. \\
  } \\
  \parbox{205pt}{\textbf{Definitions}: The function $REWRITING(x_i)$ returns a set of
   candidate sentences by applying sentence rewriting on sentence $x$;
   $PARSE(p_\theta,x)$ parses the sentence $x$ based on current parameters $\theta$,
   using agenda-based parsing; $CHOOSEORACLE(h_0)$ chooses the derivation
   with highest reward from the root of $h_0$; $CHOOSEORACLE(H_{target})$
   chooses the derivation with highest reward from a set of derivations.
   $CHOOSEORACLE({h_{target}^*})$ chooses the new sentence that results in
   derivation with highest reward.\\
  } \\
  \parbox{205pt}{\textbf{Algorithm}:\\
  \-\hspace{.4cm} $\theta_1 \gets \bf 0$,  $\theta_2 \gets \bf 0$ \\
  \-\hspace{.4cm} \textbf{for} $t = 1...T$, $i = 1...N$: \\
  \-\hspace{.8cm} $X = REWRITING(x_i)$ \\
  \-\hspace{.8cm}  \textbf{for} each ${{x'}_i} \in X:$ \\
  \-\hspace{1.2cm}  $h_0 \gets PARSE(p_\theta,{{x'}_i})$ \\
  \-\hspace{1.2cm} $d^* \gets CHOOSEORACLE(h_0)$ \\
  \-\hspace{1.2cm} $h_{target} \gets PARSE({p_\theta^{+cw}}, {{x'}_i})$ \\
  \-\hspace{.8cm} ${h_{target}^*} \gets CHOOSEORACLE(H_{target})$ \\
  \-\hspace{.8cm} ${{x'}_i^*} \gets CHOOSEORACLE({h_{target}^*})$ \\
  \-\hspace{.8cm} $\theta_2\gets\theta_2+\eta R(h_{target}^*)\sum\nolimits_{t=1}^T\delta(h_{target}^*)$ \\
  \-\hspace{.8cm} $\theta_1\gets\theta_1+\eta R(h_{target}^*)\delta(x_i,{x'}_i^*)$  \\
  } \\
  \parbox{205pt}{\textbf{Output}: Estimated parameters $\theta_1$ and $\theta_2$.
  } \\
 \hline
 \end{tabular}
 \caption{\label{algorithm-table} Our learning algorithm for parameter estimation
 from question-answer pairs.}
 \end{table}

\subsection{Features}   

As described in Section 4.3, our model uses two kinds of features.
One for the semantic parsing module – which are simply the same features described in
Berant and Liang \shortcite{TACL646}. One for the sentence rewriting module –these features are
 defined over the original sentence, the generated sentence rewritings and the final derivations:

\noindent{\bf Features for dictionary-based rewriting}.  Given a sentence $s_0$, when the new sentence $s_1$ is generated
 by replacing a word to its explanation $w\to ex$, we will generate four features: The first feature indicates
 the word replaced. The second feature indicates the replacement $w\to ex$ we used. The final two features are
 the POS tags of the left word and the right word of $w$ in $s_0$.

 \noindent{\bf Features for template-based rewriting}.  Given a sentence $s_0$, when the new sentence $s_1$ is generated
 through a template based rewriting $t_1\to t_2$, we generate four features: The first feature indicates the
 template pair ($t_1$, $t_2$) we used. The second feature is the similarity between the sentence $s_0$ and
 the template $t_1$, which is calculated using the word overlap between $s_0$ and $t_1$.
 The third feature is the compatibility of the template pair, which is the pointwise mutual information
 (PMI) between $t_1$ and $t_2$ in the WikiAnswers corpus. The final feature is triggered when the target logical
 form only contains an atomic formula (or predicate), and this feature indicates the mapping from
  template $t_2$ to the predicate $p$.

\section{Experiments}   

In this section, we assess our method and compare it with other methods.

\subsection{Experimental Settings}   

{\bf Dataset}: We evaluate all systems on the benchmark WEBQUESTIONS dataset ~\cite{berant-EtAl:2013:EMNLP},
 which contains 5,810 question-answer pairs. All questions are collected by crawling
 the Google Suggest API, and their answers are obtained using Amazon Mechanical Turk.
 This dataset covers several popular topics and its questions are commonly asked on the web.
  According to Yao \shortcite{yao:2015:demos}, 85\% of questions can be answered by predicting a single binary relation.
  In our experiments, we use the standard train-test split ~\cite{berant-EtAl:2013:EMNLP},
  i.e., 3,778 questions (65\%) for training and 2,032 questions (35\%) for testing,
  and divide the training set into 3 random 80\%-20\% splits for development.

  Furthermore, to verify the effectiveness of our method on solving the vocabulary mismatch problem,
  we manually select 50 mismatch test examples from the WEBQUESTIONS dataset, where all sentences
  have different structure with their target logical forms, e.g., ``\textit{Who is keyshia cole dad?}'' and
   ``\textit{What countries have german as the official language?}''.

 \noindent{\bf System Settings}: In our experiments, we use the Freebase Search API for entity lookup.
   We load Freebase using Virtuoso, and execute logical forms by converting them to SPARQL and
    querying using Virtuoso. We learn the parameters of our system by making three passes over
     the training dataset, with the beam size $K=200$, the dictionary rewriting size $K_D=100$,
     and the template rewriting size $K_T=100$.

\noindent {\bf Baselines}: We compare our method with several traditional systems,
 including semantic parsing based systems ~\cite{berant-EtAl:2013:EMNLP,berant-liang:2014:P14-1,TACL646,yih-EtAl:2015:ACL-IJCNLP},
 information extraction based systems ~\cite{yao-vandurme:2014:P14-1,yao:2015:demos},
 machine translation based systems ~\cite{bao-EtAl:2014:P14-1}, embedding based systems ~\cite{bordes-chopra-weston:2014:EMNLP2014,yang-EtAl:2014:EMNLP2014},
 and QA based system ~\cite{DBLP:conf/cikm/BastH15}.

\noindent {\bf Evaluation}: Following previous work ~\cite{berant-EtAl:2013:EMNLP}, we evaluate different systems
 using the fraction of correctly answered questions. Because golden answers may have multiple values,
 we use the average F1 score as the main evaluation metric.

\subsection{Experimental Results}

Table 8 provides the performance of all base-lines and our method. We can see that:

\begin{enumerate}
  \item 	Our method achieved competitive performance: Our system outperforms all baselines
  and get the best F1-measure of 53.1 on WEBQUESTIONS dataset.
  \item 	Sentence rewriting is a promising technique for semantic parsing:
  By employing sentence rewriting, our system gains a 3.4\% F1 improvement over the base system
  we used ~\cite{TACL646}.
  \item 	Compared to all baselines, our system gets the highest precision.
  This result indicates that our parser can generate more-accurate logical forms by sentence rewriting.
  Our system also achieves the third highest recall, which is a competitive performance.
  Interestingly, both the two systems with the highest recall ~\cite{DBLP:conf/cikm/BastH15,yih-EtAl:2015:ACL-IJCNLP}
  rely on extra-techniques such as entity linking and relation matching.
\end{enumerate}

\begin{table}
\small
\centering
\begin{tabular}{|c|c|c|c|}
\hline
\bf System & \bf Prec. & \bf Rec. & \bf F1 (avg) \\ \hline
Berant et al., 2013 & 48.0 & 41.3 & 35.7 \\
Yao and Van-Durme, 2014 & 51.7 & 45.8 & 33.0 \\
Berant and Liang, 2014 & 40.5  &  46.6 & 39.9 \\
Bao et al., 2014 & -- & -- & 37.5\\
Bordes et al., 2014a & -- & -- & 39.2 \\
Yang et al., 2014 & --  & -- & 41.3\\
Bast and Haussmann, 2015 & 49.8 & 60.4  & 49.4 \\
Yao, 2015 & 52.6 & 54.5 & 44.3 \\
Berant and Liang, 2015 & 50.5 & 55.7 & 49.7\\
Yih et al., 2015 & 52.8 & \textbf{60.7} & 52.5\\ \hline
Our approach & \textbf{53.7} & 60.0 & \textbf{53.1}\\
\hline
\end{tabular}
\caption{\label{result-table}The results of our system and recently published systems.
The results of other systems are from either original papers or the standard evaluation
web.}
\end{table}

\noindent{\bf The effectiveness on mismatch problem}. To analyze the commonness of mismatch problem in semantic parsing,
we randomly sample 500 questions from the training data and do manually analysis, we found that 12.2\%
out of the sampled questions have mismatch problems: 3.8\% out of them have 1-N mismatch problem
and 8.4\% out of them have N-1 mismatch problem.

To verify the effectiveness of our method on solving the mismatch problem,
we conduct experiments on the 50 mismatch test examples and Table 9 shows the performance.
We can see that our system can effectively resolve the mismatch between natural language and
 target ontology: compared to the base system, our system achieves a significant 54.5\% F1 im-provement.

 \begin{table}[h]
 \centering
 \small
 \begin{tabular}{|p{85pt}|p{25pt}|p{25pt}|p{35pt}|}
 \hline
 \parbox{85pt}{\centering\textbf{System}
 } & \parbox{25pt}{\centering\textbf{Prec.}
 } & \parbox{25pt}{\centering\textbf{Rec.}
 } & \parbox{35pt}{\centering\textbf{F1 (avg)}
 } \\ \hline
 \parbox{85pt}{\centering{Base system}
 } & \parbox{25pt}{\centering{31.4}
 } & \parbox{25pt}{\centering{43.9}
 } & \parbox{35pt}{\centering{29.4}
 } \\ \hline
 \parbox{85pt}{\centering{Our system}
 } & \parbox{25pt}{\centering\textbf{83.3}
 } & \parbox{25pt}{\centering\textbf{92.3}
 } & \parbox{35pt}{\centering\textbf{83.9}
 } \\
 \hline
 \end{tabular}
 \caption{\label{result-50-table}The results on the 50 mismatch test dataset.}
 \end{table}

 When scaling a semantic parser to open-domain situation or web situation,
 the mismatch problem will be more common as the ontology and language complexity increases
 ~\cite{kwiatkowski-EtAl:2013:EMNLP}. Therefore we believe the sentence rewriting method proposed in this paper
 is an important technique for the scalability of semantic parser.

\noindent{\bf The effect of different rewriting algorithms}. To analyze the contribution of different rewriting
methods, we perform experiments using different sentence rewriting methods and the results are
 presented in Table 10. We can see that:

 \begin{table}
 \centering
 \small
 \begin{tabular}{|p{85pt}|p{25pt}|p{25pt}|p{35pt}|}
 \hline
 \parbox{85pt}{\centering\textbf{Method}
 } & \parbox{25pt}{\centering\textbf{Prec.}
 } & \parbox{25pt}{\centering\textbf{Rec.}
 } & \parbox{35pt}{\centering\textbf{F1 (avg)}
 } \\ \hline
 \parbox{85pt}{\centering{base}
 } & \parbox{25pt}{\centering{49.8}
 } & \parbox{25pt}{\centering{55.3}
 } & \parbox{35pt}{\centering{49.1}
 } \\ \hline
 \parbox{85pt}{\centering{+ dictionary SR (only)}
 } & \parbox{25pt}{\centering{51.6}
 } & \parbox{25pt}{\centering{57.5}
 } & \parbox{35pt}{\centering{50.9}
 } \\ \hline
 \parbox{85pt}{\centering{+ template SR (only)}
 } & \parbox{25pt}{\centering{52.9}
 } & \parbox{25pt}{\centering{59.0}
 } & \parbox{35pt}{\centering{52.3}
 } \\ \hline
 \parbox{85pt}{\centering{+ both}
 } & \parbox{25pt}{\centering\textbf{53.7}
 } & \parbox{25pt}{\centering\textbf{60.0}
 } & \parbox{35pt}{\centering\textbf{53.1}
 } \\
 \hline
 \end{tabular}
 \caption{\label{result-2032-table}The results of the base system and our systems on the 2032 test questions.}
 \end{table}

 \begin{enumerate}
   \item Both sentence rewriting methods improved the parsing performance, they resulted in 1.8\%
   and 3.2\% F1 improvements respectively\footnote{Our base system yields a slight drop
   in accuracy compared to the original system (Berant and Liang, 2015),
   as we parallelize the learning algorithm, and the order of the data for updating the parameter
   is different to theirs.}.
   \item 	Compared with the dictionary-based rewriting method, the template-based rewriting method
   can achieve higher performance improvement. We believe this is because N-1 mismatch problem is more
    common in the WEBQUESTIONS dataset.
   \item 	The two rewriting methods are good complementary of each other. The semantic parser can
   achieve a higher performance improvement when using these two rewriting methods together.
  \end{enumerate}

  \noindent{\bf The effect on improving robustness}. We found that the template-based rewriting method
  can greatly improve the robustness of the base semantic parser. Specially, the template-based method
  can rewrite similar sentences into a uniform template, and the (template, predicate) feature
   can provide additional information to reduce the uncertainty during parsing.
   For example, using only the uncertain alignments from the words ``\textit{people}'' and ``\textit{speak}'' to
    the two predicates \texttt{official\_language} and \texttt{language\_spoken}, the base parser will parse the sentence
    ``\textit{What does jamaican people speak?}''
    into the incorrect logical form \texttt{official\_language.jamaican} in our experiments, rather than into the correct form
    \texttt{language\_spoken.jamaican} (See the final example in Table 11). By exploiting the alignment from the template
    ``\textit{what language does \$y people speak}'' to the predicate  ,
    our system can parse the above sentence correctly.

    \begin{table}
    \small
    \centering
    \begin{tabular}{|p{10pt}|p{185pt}|}
      \hline
      \parbox{10pt}{\textbf{O}
      } & \parbox{185pt}{\textit{Who is willow smith mom name?}
      } \\ \hline
      \parbox{10pt}{\textbf{R}
      } & \parbox{185pt}{\textit{Who is willow smith female parent name?}
      } \\ \hline
      \parbox{10pt}{\textbf{LF}
      } & \parbox{185pt}{\texttt{parentOf.willow\_smith$\wedge$gender.female}
      } \\ \hline \hline
      \parbox{10pt}{\textbf{O}
      } & \parbox{185pt}{\textit{Who was king henry viii son?}
      } \\ \hline
      \parbox{10pt}{\textbf{R}
      } & \parbox{185pt}{\textit{Who was king henry viii male child?}
      } \\ \hline
      \parbox{10pt}{\textbf{LF}
      } & \parbox{185pt}{\texttt{childOf.king\_henry$\wedge$gender.male}
      } \\ \hline \hline
      \parbox{10pt}{\textbf{O}
      } & \parbox{185pt}{\textit{What are some of the traditions of islam?}
      } \\ \hline
      \parbox{10pt}{\textbf{R}
      } & \parbox{185pt}{\textit{What is of the religion of islam?}
      } \\ \hline
      \parbox{10pt}{\textbf{LF}
      } & \parbox{185pt}{\texttt{religionOf.islam}
      } \\ \hline \hline
      \parbox{10pt}{\textbf{O}
      } & \parbox{185pt}{\textit{What does jamaican people speak?}
      } \\ \hline
      \parbox{10pt}{\textbf{R}
      } & \parbox{185pt}{\textit{What language does jamaican people speak?}
      } \\ \hline
      \parbox{10pt}{\textbf{LF}
      } & \parbox{185pt}{\texttt{language\_spoken.jamaica}
      } \\
      \hline
    \end{tabular}
    \caption{\label{examples-table}Examples which our system generates more accurate logical
    form than the base semantic parser. \textbf{O} is the original sentence;
    \textbf{R} is the generated sentence from sentence rewriting (with the highest score for the model,
    including rewriting part and parsing part); \textbf{LF} is the target logical form.}
    \end{table}

  \noindent{\bf The effect on OOV problem}. We found that the sentence rewriting method can also
  provide extra profit for solving the OOV problem. Traditionally, if a sentence contains a word which is not covered
  by the lexicon, it will cannot be correctly parsed. However, with the help of sentence rewriting,
  we may rewrite the OOV words into the words which are covered by our lexicons. For example,
   in Table 11 the 3rd question ``\textit{What are some of the traditions of islam?}'' cannot be correctly parsed
   as the lexicons don’t cover the word ``\textit{tradition}''. Through sentence rewriting, we can generate a new
   sentence ``\textit{What is of the religion of islam?}'', where all words are covered by the lexicons,
   in this way the sentence can be correctly parsed.

\subsection{Error Analysis}

To better understand our system, we conduct error analysis on the parse results.
Specifically, we randomly choose 100 questions which are not correctly answered by our system.
We found that the errors are mainly raised by following four reasons (See Table 12 for detail):

\begin{table}[h]
\small
\centering
\begin{tabular}{|p{45pt}|p{27pt}|p{110pt}|}
  \hline
  \parbox{45pt}{\centering\textbf{Reason}
  } & \parbox{27pt}{\centering\textbf{\#(Ratio)}
  } & \parbox{110pt}{\centering\textbf{Sample Example}
 } \\
 \hline
 \parbox{45pt}{\centering Label issue
 } & \parbox{27pt}{\centering 38
 } & \parbox{110pt}{\centering\textit{What band was george clinton in?}
} \\
\hline
\parbox{45pt}{\centering N-ary predi-cate($n>2$)
} & \parbox{27pt}{\centering 31
} & \parbox{110pt}{\centering\textit{What year did the seahawks win the superbowl?}
} \\
\hline
\parbox{45pt}{\centering Temporal clause
} & \parbox{27pt}{\centering 15
} & \parbox{110pt}{\centering\textit{Who was the leader of the us during wwii?}
} \\
\hline
\parbox{45pt}{\centering Superlative
} & \parbox{27pt}{\centering 8
} & \parbox{110pt}{\centering\textit{Who was the first governor of colonial south carolina?}
} \\
\hline
\parbox{45pt}{\centering Others
} & \parbox{27pt}{\centering 8
} & \parbox{110pt}{\centering\textit{What is arkansas state capitol?}
} \\
\hline
\end{tabular}
\caption{\label{error-table}The main reasons of parsing errors,
 the ratio and an example for each reason are also provided.}
\end{table}

The first reason is the label issue. The main label issue is incompleteness, i.e.,
 the answers of a question may not be labeled completely. For example,
 for the question ``\textit{Who does nolan ryan play for?}'', our system returns 4 correct teams but
 the golden answer only contain 2 teams. One another label issue is the error labels. For example,
 the gold answer of the question ``\textit{What state is barack obama from?}'' is labeled as ``\textit{Illinois}'',
 however, the correct answer is ``\textit{Hawaii}''.

 The second reason is the n-ary predicate problem ($n>2$). Currently, it is hard for a parser
 to conduct the correct logical form of n-ary predicates. For example, the question
 ``\textit{What year did the seahawks win the superbowl?}'' describes an n-ary championship event,
 which gives the championship and the champion of the event, and expects the season.
 We believe that more research attentions should be given on complicated cases,
 such as the n-ary predicates parsing.

 The third reason is temporal clause. For example, the question ``\textit{Who did nasri play for before arsenal?}''
 contains a temporal clause ``\textit{before}''. We found temporal clause is complicated and makes it strenuous
 for the parser to understand the sentence.

 The fourth reason is superlative case, which is a hard problem in semantic parsing.
  For example, to answer ``\textit{What was the name of henry viii first wife?}'',
  we should choose the first one from a list ordering by time. Unfortunately,
  it is difficult for the current parser to decide what to be ordered and how to order.

There are also many other miscellaneous error cases,
such as spelling error in the question, e.g.,
``\textit{capitol}'' for ``\textit{capital}'', ``\textit{mary}'' for ``\textit{marry}''.

\section{Conclusions}

In this paper, we present a novel semantic parsing method,
which can effectively deal with the mismatch between natural language
and target ontology using sentence rewriting.
We resolve two common types of mismatch (i) one word in natural language sentence
 vs one compound formula in target ontology (1-N),
 (ii) one complicated expression in natural language sentence
  vs one formula in target ontology (N-1). Then we present two sentence rewriting methods,
   dictionary-based method for 1-N mismatch and template-based method for N-1 mismatch.
   The resulting system significantly outperforms the base system on the WEBQUESTIONS dataset.

Currently, our approach only leverages simple sentence rewriting methods.
In future work, we will explore more advanced sentence rewriting methods.
Furthermore, we also want to employ sentence rewriting techniques for
other challenges in semantic parsing, such as the spontaneous,
 unedited natural language input, etc.

\bibliographystyle{acl2016}
\bibliography{acl2016}

\end{document}